\documentclass[accepted]{uai2026} 
                        
\usepackage[american]{babel}
\usepackage{csquotes}
\usepackage{booktabs} 
\usepackage{tikz} 

\hypersetup{pdfborder={0 0 0}}

\usepackage{algorithm}
\usepackage{algpseudocode} 
\usepackage{setspace} 

\usepackage[
    backend=biber,
    style=apa,
    isbn=false,
    url=false
]{biblatex}  

\AtEveryBibitem{%
  \clearfield{issn}%
}

\usepackage{amsmath}
\usepackage{amssymb}
\usepackage{mathtools}
\usepackage{amsthm}

\usetikzlibrary{calc}
\usepackage{tikz}
\usepackage{tikz-cd}
\usetikzlibrary{arrows.meta}

\newcommand{\citecolor}[1]{\textcolor{darkgray!90}{#1}}
\renewcommand{\cite}[1]{\citecolor{\parencite{#1}}}
\newcommand{\citet}[1]{\citecolor{\textcite{#1}}}

\newcommand*{\scriptA}{\ensuremath{\mathcal{A}}}
\newcommand*{\scriptB}{\ensuremath{\mathcal{B}}}

\newcommand*{\scriptD}{\ensuremath{\mathcal{D}}}

\newcommand*{\scriptG}{\ensuremath{\mathcal{G}}}

\newcommand*{\scriptJ}{\ensuremath{\mathcal{J}}}

\newcommand*{\scriptL}{\ensuremath{\mathcal{L}}}

\newcommand*{\scriptN}{\ensuremath{\mathcal{N}}}
\newcommand*{\scriptO}{\ensuremath{\mathcal{O}}}

\newcommand*{\scriptS}{\ensuremath{\mathcal{S}}}

\newcommand*{\scriptZ}{\ensuremath{\mathcal{Z}}}

\newcommand*{\doubleE}{\ensuremath{\mathbb{E}}}

\newcommand*{\doubleP}{\ensuremath{\mathbb{P}}}

\newcommand*{\doubleR}{\ensuremath{\mathbb{R}}}

\newcommand{\kldiv}{\mathit{KL}}

\newcommand{\indep}{\perp \!\!\! \perp}

\DeclareMathOperator*{\argmax}{arg\,max}

\newcommand{\myeqref}[1]{Eq.~(\ref{#1})}

\theoremstyle{plain}
\newtheorem{theorem}{Theorem}[section]
\newtheorem{proposition}[theorem]{Proposition}

\newtheorem{corollary}[theorem]{Corollary}

\theoremstyle{definition}

\theoremstyle{remark}

\newenvironment{numproof}[1][Proof]{%
  \refstepcounter{proofnum} 
  \par\noindent\textbf{#1 \theproofnum. }\ignorespaces
}{%
  \hfill$\qed$\par
}

\title{VariBASed: Variational Bayes-Adaptive\\Sequential Monte-Carlo Planning for Deep Reinforcement Learning}

\author[1,2]{\href{mailto:<joery@trent.ai>?Subject=VariBASeD}{Joery A. de Vries}{}}
\author[2]{Jinke He}
\author[2]{Yaniv Oren}
\author[2]{Pascal R. van der Vaart}
\author[2]{Mathijs M. de Weerdt}
\author[2]{Matthijs T. J. Spaan }
\affil[1]{%
    Trent AI Ltd\\
    London, United Kingdom
}
\affil[2]{%
    EEMCS\\
    Delft University of Technology\\
    Delft, the Netherlands
}
  
\begin{document}
\maketitle

\begin{abstract}
  Optimally trading-off exploration and exploitation is the holy grail of reinforcement learning as it promises maximal data-efficiency for solving any task.
  Bayes-optimal agents achieve this, but obtaining the belief-state and performing planning are both typically intractable. 
  Although deep learning methods can greatly help in scaling this computation, existing methods are still costly to train.
  To accelerate this, this paper proposes a variational framework for learning and planning in Bayes-adaptive Markov decision processes that coalesces variational belief learning, sequential Monte-Carlo planning, and meta-reinforcement learning.
  In a single-GPU setup, our new method VariBASeD exhibits favorable scaling to larger planning budgets, improving sample- and runtime-efficiency over prior methods.
\end{abstract}

\section{Introduction}

Reinforcement Learning ({rl}) describes how to find an optimal policy for sequential decision making by maximizing expected returns under uncertainty \cite{sutton_reinforcement_2018}.
In recent years, this framework has enjoyed many successes such as achieving superhuman performance in boardgames \cite{silver_2018_alphazero}, and displaying rapid adaptation to test tasks \cite{bauer_humantimescale_2023}.
The crux of this methodology is to optimally trade off gathering informative data (exploration) with focusing on data that guarantee high expected returns (exploitation).
This is relevant in risk sensitive applications, since exploration and exploitation in an inaccurate model of the world often incurs significant repercussions.

A Bayes-optimal agent achieves an optimal balance of exploration and exploitation \cite{duff_optimal_2002}.
It does so by augmenting its state-space with a belief over hypothetical environments given past data (meaning the current trajectory's history).
It then selects its actions by accounting for the expected returns but also on how its own uncertainty evolves (the belief state).
Unfortunately, computing this agent is hopelessly intractable due to the many nested integrals one has to deal with, this includes the marginalization over prior reward and transition functions at every new datapoint.
Rightfully so, this has motivated development of heuristics like posterior sampling \cite{osband_psrl_2013} or optimism \cite{munos_optimism_2014}.
However, such heuristics typically sacrifice optimizing a ``true objective'' for the sake of tractability.

Instead, our focus is on direct approximations of the Bayes-optimal agent.
Recently, variational approximations \cite{bishop2007} combined with amortized probabilistic inference \cite{margossian2024amortized} using deep neural networks have shown great promise in scaling to high dimensional problem settings.
For instance, this has been successfully tried for approximating beliefs \cite{mnih_neural_2014, igl_deepvar_2018, becker2024kalmambaefficientprobabilisticstate}, policy inference \cite{haarnoja_sac_2018, piche2018probabilistic}, and Bayes-optimal agents \cite{zintgraf_varibad_2020, iqbal2024nestingparticlefiltersexperimental}.

A key element in developing general {rl} algorithms, including approximate Bayes-optimal algorithms, that scale to higher dimensions, larger datasets, and greater (distributed) compute resources is careful engineering to match the intended hardware. 
This aspect is often less discussed but plays a subtle and essential role.
This is especially relevant for approximating Bayes-optimal agents due to their history dependence.
In deep learning, history dependence is typically tackled with recurrent neural network (RNN) architectures which require unrolling and BackPropagation Through Time (BPTT) for training with stochastic gradient descent.
Unfortunately, this unrolling becomes expensive in memory for long sequences, is difficult to parallelize, and is also often difficult to train reliably \cite{goodfellow_deep_2016}. 
Furthermore, they often require vast amounts of data and large model sizes in order to find a decent policy \cite{bauer_humantimescale_2023}. 
Our Bayes-optimal control problem fits within this description.
Bayes optimal control requires a single agent to infer the true MDP, solve (possibly infinitely) many Markov decision processes (MDP; \citeauthor{puterman_markov_1994}, \citeyear{puterman_markov_1994}), and anticipate consequences of exploratory actions; leading to a much more difficult learning task compared to solving a single MDP.
Thus, pushing the frontiers of current methodologies necessitates the development of frameworks that increase the scaling capabilities of throughput, whether it is in computation, data, or memory.
Fortunately, recent frameworks like PureJaxRL \cite{lu2022discovered} help in scaling up throughput of training by dispatching large computations on GPUs which excel at highly parallelized data processing.

Transformer architectures \cite{vaswani2017attention} have recently been established as alternatives to RNNs, often showing much stronger performance while reducing training time due to improved parallelization over recurrent BPTT.
Unfortunately, a key limitation of efficient GPU based frameworks is their strict adherence to computations on static shapes, whereas many history dependent control problems vary in input length.
This is a strong restriction for utilizing more scalable and performant architectures such as Transformers \cite{parisotto2021efficient, vaswani2017attention}. 
Unlike RNNs, Transformers do not maintain an explicit state mechanism that can be cached throughout fixed size computations, they strictly require dynamically sized memories which cause overhead in GPU dispatch for resizing.
Alternatively, they require padding of the history with empty observations, which also introduces unnecessary overhead. 
Interestingly, recent theoretical work \cite{katharopoulos_transformers_2020, gu2024mamba} shows that Transformers can be viewed as a form of state-space models. 
This perspective suggests that linear state dynamics (of the form $x_t = A x_{t-1} + B u_t + C$) are the key ingredient behind scalable training on very long sequences, thanks to the associativity of linear recurrences.
Furthermore, recent developments in state-space sequence models for RL \cite{lu_s5rl_2023, smith2023simplifiedstatespacelayers, becker2024kalmambaefficientprobabilisticstate} have drastically improved their performance compared to previously popular recurrent neural network \cite{duan_rl2_2016} or Transformer based approaches \cite{parisotto2021efficient}, at much improved training and inference computation time.
For instance, state-space models can perform efficient inference in $\mathcal{O}(1)$ time, yet they are also trainable in $\mathcal{O}(\log n)$ time ($n$ being the sequence length) by using GPU parallelization.
Instead, recurrent network training scales with $\mathcal{O}(n)$ time and Transformer inference with $\mathcal{O}(n^2)$ \cite{vaswani2017attention} or $\mathcal{O}(n)$ with KV-caching \cite{kvcache_pope_2023}.

Building on these considerations about computational efficiency, we next consider a complementary strategy with model-based RL.
In MDP settings, a consensus is that model-based RL \cite{sutton_reinforcement_2018, hafner2020dream} improves data-efficiency (sample-efficiency) of algorithms by spending on average more compute per datapoint compared to model-free approaches.
However, prior work has also shown that model-based RL utilizing online {smc} planning can improve \textit{algorithm runtime} compared to model-free RL \cite{macfarlane2024spo, vries2025trustregion} when a model of the environment is available.
This is specifically relevant when environment interactions are also comparatively cheap to model training (gradient descent), since online planning often improves the quality of training data and learning targets \cite{hamrick2021roleofplanning}.
In typical MDPs this bottleneck depends strongly on the environment and the agent's model size, however, when specifically dealing with history dependent policies, this bottleneck can quickly shift in favor of taking more environment samples.
The reason for this is that training on long trajectories at least requires a $O(\log n)$ time complexity for state-space models, whereas increasing the planning budget at decision time only amounts to a bigger constant in the inference complexity\footnote{Naturally, the planning budget (which induces the constant) should be adjusted based on the length of the trajectory. However, the optimal trade-off between planning budget, unroll length (or SGD budget) is a question we defer for future work.}.

Towards improving the scalability of training Bayes-adaptive agents, this paper introduces a new algorithm which we dub VariBASeD.
VariBASeD combines a sequential Monte-Carlo planner in combination with amortized variational inference \cite{agrawal_amortized_2021} to reliably train a Bayes adaptive agent.
In essence, we phrase our Bayes adaptive agent as an intractable \emph{target} distribution from which we draw approximate samples using more manageable \emph{proposal} distributions that we design through efficient particle-filter methods \cite{chopin_introduction_2020}.
We show that online {smc} planners can improve both the sample- and runtime-efficiency for training agents that approximate Bayes-optimal exploration and exploitation when compared to model-free methods like RL$^2$ \cite{duan_rl2_2016}. 
In summary our contributions are:
\begin{enumerate}
    \item \textbf{Theoretical Framework:} We derive a history-dependent Expectation-Maximization (EM) method for approximate policy iteration with learned (amortized) belief states.
    \item \textbf{Algorithmic Integration:} We develop a history-dependent importance-sampling strategy that integrates SMC-planning with the learned beliefs.
    \item \textbf{Hardware-Efficient Implementation:} We design a specialized single-GPU training setup that handles episode boundaries via fixed-size batch sampling, enabling high-throughput GPU training with consumer-grade hardware.
\end{enumerate}
To efficiently simulate downstream usage of Bayes-adaptive agents, we evaluated our contributions in a zero-shot multitask RL setting \cite{kirk_survey_2023}.
In this setting, we want the agent to maximize expected return on an unknown distribution over MDPs over \textit{multiple} adaptation episodes \cite{duan_rl2_2016}.
In other words, the agent's sequence model and value targets are accumulated for multiple ``inner-episodes'', they are only reset when the ``true'' underlying task is also reset.
We evaluate our method on a continuous function optimization problem and a discrete gridworld environment with uncertain rewards, and observe improved sample-efficiency and runtime as we scale up the planner's compute budget.

\section{Background}

We want to find an optimal policy $\pi$ for a sequential decision-making problem for which there is uncertainty on the reward or transition function.
Our formalization relies on an infinite-horizon Markov decision process (MDP) \cite{puterman_markov_1994}, which we write as a tuple $M=\langle \scriptS, \scriptA, p_R, p_S, \rho \rangle$.
We have states $S\in \scriptS$ and actions $A \in \scriptA$, which generate rewards $p_R(R_t | S_t, A_t) \in \doubleP(\doubleR)$ and transitions $p_S(S_{t+1} | S_t, A_t) \in \doubleP(\scriptS)$ with initial state distribution $S_1 \sim \rho(S_1)$.
Together, we write the realization for a finite sequence as $h_{1:T} = \{a_t, r_t, s_{t+1}\}^T_{t=1}$ and assume that $h_{1:\infty}$ always ends up in an absorbing state with zero rewards (this is a weak assumption that we satisfy by discounting the returns that act as a termination probability $1-\gamma \in [0, 1]$).
Our aim is to find a policy $\pi$ that satisfies, $\max_\pi \scriptJ_M(\pi) = \max_\pi \doubleE_{\pi, p_R, p_S, \rho} [\sum_{t=1}^\infty R_t]$.

\paragraph{Bayesian Reinforcement Learning.}

In our setting we assume uncertainty on the true MDP $M_{\textit{true}} \sim p(M)$, particularly we are uncertain about the true reward and state dynamics $p_R, p_S$.
This creates a non-Markovian RL problem due to induced history dependence for inferring $M$.
Bayesian RL methods \cite{duff_optimal_2002, ghavamzadeh_bayesian_2015} offer a useful framework to deal with history dependence by maintaining a posterior of MDPs over time $p(M_t | H_{<t})$.
We write such a density for a realization as,
\begin{align} \label{vb:eq:rl_history_latents}
    p_\pi(H_{1:T}, M_{1:T}) &= 
    \doubleE_{\rho} \Big[\prod_{t=1}^T p(S_{t+1}, R_t | A_{t}, S_t, M_t)\nonumber \\ &\cdot\:\pi(A_t | S_t, M_t) \cdot p(M_t | H_{<t}) \Big],
\end{align}
where $\rho = \rho(S_1 | M_0)p(M_0)$ factorizes as some initial state density with a known MDP-prior, and we combined the independent terms $S_{t+1}, R_t$ for brevity.

In essence, the above density imposes structure to the marginal distribution for $p_\pi(H_{1:T})$ by augmenting the dynamics with an evolving posterior distribution over MDPs.
For this reason, the posterior is often summarized in terms of a belief state $b_t = p(M_t | H_{<t})$ that augments the dynamics of the base MDP.
We can write this new MDP as $M^+ = \langle \scriptS^+, \scriptA, p_{R}^+, p_{S}^+, \rho^+ \rangle$, where $\scriptS^+ = \scriptS \times \scriptB$ is the belief $b_t \in \scriptB$ augmented state-space, such that the dynamics, rewards, and policy satisfy,
\begin{align}
    p_{S}^+(b_{t+1}, S_{t+1} | S_t, A_t, b_t) &= 
    \delta_{b_{t+1}} \doubleE_{b_t}p(S_{t+1} | A_t, S_t, M_t), \nonumber
    \\[6pt] 
    p_{R}^+(R_{t} | S_t, A_t, b_t) &= \doubleE_{b_t}p(R_t | A_t, S_t, M_t), \nonumber
    \\
    \pi^+(A_{t} | S_t, b_t) &= \doubleE_{b_t}\pi(A_t | S_t, M_t). \nonumber
\end{align}

The augmented MDP $M^+$ is also known as a Bayes-adaptive MDP (BA-MDP), for which the optimal policy $\pi^+:~\scriptS\times\scriptB\rightarrow\doubleP(\scriptA)$ is a \textit{Bayes-optimal policy} \cite{duff_optimal_2002}.
In other words, a policy that optimally trades-off exploration and exploitation.
This differs from common heuristics like posterior sampling or optimism \cite{osband_psrl_2013, osband_why_2017} since $\pi^+$ explicitly accounts for how the complete belief evolves given additional data and its impact on expected returns.

\paragraph{Belief Amortization through Meta-Learning.}

Finding $\pi$ in the augmented state-space $\scriptS \times \scriptB$ is, unfortunately, intractable for most interesting problems.
Furthermore, tracking the belief state $b_t$ itself is usually also intractable, assumes knowledge of the density for the reward and state dynamics, and requires specifying a reasonable prior $b_0$.
To deal with this intractability, a successful approach is to use amortization \cite{amos_tutorial_2023} in combination with flexible function approximation.
For this, we assume that we can approximate the true belief, reward, and state dynamics, and the policy, with e.g., a large neural network $h_\theta(\cdot)$.
Then we perform inference to the probabilistic graphical model of \myeqref{vb:eq:rl_history_latents} by fitting the neural network parameters $\theta$ directly on (a proxy of) the evidence $p_\pi(H_{1:T})$.
In other words, amortization means that we can rely on comparatively cheap neural network evaluations in downstream tasks that approximate probabilistic inference to our (intractable) model after performing an expensive training procedure.

A flexible setup for such a training procedure is through meta learning \cite{mikulik_metatrained_2020}, which only requires us to be able to sample environments from a distinct generator of MDPs $\scriptG_{\textit{train}}(M)$, not to be confused with the MDP prior $p(M)$.
For brevity, let us write the joint density of \myeqref{vb:eq:rl_history_latents} as $p_\pi(H_{1:t}, M_{1:t} | \theta)$ so that each component now depends on neural network parameters $\theta$.
We will denote the belief $b_t$ as a parametric family $b_{\phi_t} = p(M | \phi_t = h_\theta(H_{<t}))$, where our neural network $h_\theta$ maps histories to the belief parameters $\phi_t$.
The idea is to compute the policy objective $\hat{\scriptJ}_M(\theta) = \doubleE_{p_\pi(H_{1:\infty} \mid \theta)}[\sum_{t=1}^\infty R_t]$ in expectation of the environment generator, and perform stochastic gradient ascent on the neural network parameters for $h_\theta$.
This assumes end-to-end differentiability for $\theta$, where we can write the optimization objective as the l.h.s. of the lower-bound,
\begin{align}  \label{vb:eq:amortization_lowerbound}
    \max_\theta 
    \doubleE_{M \sim \scriptG_{\textit{train}}} [\hat{\scriptJ}_{M}(\theta)] \le \doubleE_{M \sim \scriptG_{\textit{train}}} [\max_\pi \scriptJ_{M}(\pi)],
\end{align}
which can, e.g., be optimized through Monte-Carlo combined with REINFORCE \cite{Williams1992REINFORCE} and backpropagation through time \cite{duan_rl2_2016}.
Effectively, fitting $\theta$ in the l.h.s. of \myeqref{vb:eq:amortization_lowerbound} distills the uncertainty induced by the true rewards and transitions of sample MDPs $M \sim \scriptG_{\textit{train}}$ into the predicted belief parameters $\phi_t, \forall t$. 
Thus, we obtain a Bayes-optimal policy (at $\theta^*$) for a prior with an inductive bias \cite{finn2017modelagnostic} to the MDPs following $\scriptG_{\textit{train}}$.

When we directly model the density $p_\pi(H_{1:t}, M_{1:t} | \theta)$, we make the assumption that the posterior over MDPs $b_{\phi_t} = p_\pi(M_{t} | h_\theta(H_{<t}))$ can be captured by our model.
Even when our neural networks are sufficiently expressive, it often helps to add structure to the modeling problem.
In this regard, variational approximations \cite{kingma2022autoencodingvariationalbayes, zintgraf_varibad_2020, igl_deepvar_2018, mnih_neural_2014} can greatly help as they only assume that we can model a simplified posterior $b^q_t \approx b_t$, or with neural networks $b_{\phi_t}^q = q_\theta(M_t | H_{<t}) \approx p(M_t | H_{<t})$.
For variational distributions $q_\theta$ it is easier to assume end-to-end differentiability due to simplifying assumptions on the parametric family, e.g., enabling reparameterized gradients \cite{kingma2022autoencodingvariationalbayes}.
The variational parameters for $b^q_{\phi_t}$ are then fitted by augmenting the objective $\scriptJ_M$ with a regularized decoding loss.
Our method also adopts this variational Bayes framework for the MDP posterior, as detailed in Section~\ref{vb:sec:varbelief_learning}.

Finally, an important distinction between meta-RL and conventional RL is that the augmented dynamics for termination and episodic learning are handled differently \cite{duan_rl2_2016, zintgraf_varibad_2020}.
In particular, if a sequence of states $S_t$ ends up in an absorbing state, which typically signals that the environment can be reset to a new initial state (i.e., a new episode), then this should not reset the belief state.
The belief states are only reset when the true MDP changes.
This means that episodic learning becomes an essential part of the augmented dynamics: both the returns from future episodes and data from previous episodes are taken into account when optimizing \myeqref{vb:eq:amortization_lowerbound}.

\paragraph{Expectation-Maximization.}
We will now consider optimizing the l.h.s. of \myeqref{vb:eq:amortization_lowerbound} using Expectation-Maximization (EM) \cite{Neal1998} methods by leveraging the Control-as-Inference framework \cite{levine2018reinforcement, ziebart_modeling_2010}.
These methods embed the control aspect as part of the probabilistic graphical model, which gives rise to a flexible energy-based formulation of the RL problem.
This is in contrast to typical approaches, which treat the policy as separate from the environment \cite{puterman_markov_1994}.
Most importantly, this framework naturally enables regularized approximate policy iteration \cite{geist_theory_2019}, which has proven extremely effective \cite{abdolmaleki2018maximum, macfarlane2024spo} compared to unregularized methods such as plain REINFORCE.

\begin{figure}[t]
    \centering
    \makebox[\linewidth][c]{%
        \resizebox{0.38\linewidth}{!}{\begin{tikzpicture}

    \draw (2.15, -.75) -- (5.35, -0.75);
    \draw (2.15, -.75) -- (2.15, 3.75);
    \draw (5.35, -.75) -- (5.35, 3.75);
    \draw (2.15, 3.75) -- (5.35, 3.75);
    
    \node[minimum size=1cm] (s1) at (1.45, 0) {};
    \node[circle, draw, double, double distance=0.3mm, minimum size=1cm] (s2) at (3, 0) {$S_t, b_t$};
    \node [minimum size=.5cm] (s3) at (6, -0) {$\dots$};
    
    \node[minimum size=1cm] (a1) at (1.5, 1.5) {};
    \node[circle, draw, minimum size=1cm] (a2) at (4.5, 1.5) {$A_t$};

    \node[] (r1) at (1, 3) {};
    \node[draw, rectangle, minimum size=.8cm] (r2) at (3, 3) {$R_t$};
    
    \node[circle, draw, minimum size=1cm] (o2) at (4.5, 3) {$\scriptO_t$};
    
    \node (tt) at (4.4, -0.55) {$t=0,\dots,T$};

    \draw[-{Triangle[length=2mm, fill=black, width=1.5mm, line width=0.3mm]}] (2, 0) -- (s2);
    \draw[-{Triangle[length=2mm, fill=black, width=1.5mm, line width=0.3mm]}] (s2) -- (s3);

    \draw[-{Triangle[length=2mm, fill=black, width=1.5mm, line width=0.3mm]}] (s2) -- (a2);

    \draw[-{Triangle[length=2mm, fill=black, width=1.5mm, line width=0.3mm]}] (2, 1) -- (s2);
    \draw[-{Triangle[length=2mm, fill=black, width=1.5mm, line width=0.3mm]}] (a2) -- (s3);

    \draw[-{Triangle[length=2mm, fill=black, width=1.5mm, line width=0.3mm]}] (a2) -- (r2);
    
    \draw[-{Triangle[length=2mm, fill=black, width=1.5mm, line width=0.3mm]}] (r2) -- (o2);
    
    \draw[-{Triangle[length=2mm, fill=black, width=1.5mm, line width=0.3mm]}] (r2) .. controls (3.75, 0.85) .. (s3);
    
    \draw[-{Triangle[length=2mm, fill=black, width=1.5mm, line width=0.3mm]}] (2, 0.5) -- (s2);
    
    \draw[-{Triangle[length=2mm, fill=black, width=1.5mm, line width=0.3mm]}] (s2) -- ($(r2.south)$);

\end{tikzpicture}}%
        \hspace{-0.5cm}
        \resizebox{0.38\linewidth}{!}{\begin{tikzpicture}

    
    \draw (2.15, -.75) -- (5.35, -0.75);
    \draw (2.15, -.75) -- (2.15, 3.75);
    \draw (5.35, -.75) -- (5.35, 3.75);
    \draw (2.15, 3.75) -- (5.35, 3.75);
    
    \node[minimum size=1cm] (s1) at (1.45, 0) {};
    \node[circle, draw, double, double distance=0.3mm, minimum size=1cm] (s2) at (3, 0) {$S_t, b_t$};
    \node [minimum size=.5cm] (s3) at (6, -0) {$\dots$};
    
    \node[minimum size=1cm] (a1) at (1.5, 1.5) {};
    \node[circle, draw, minimum size=1cm] (a2) at (4.5, 1.5) {$A_t$};

    \node[] (r1) at (2, 3) {};
    \node[draw, rectangle, minimum size=.8cm] (r2) at (3, 3) {$R_t$};
    
    \node[circle, draw, fill=gray!30, minimum size=1cm] (o2) at (4.5, 3) {$\scriptO_{t:T}$};
    
    \node (tt) at (4.4, -0.55) {$t=0,\dots,T$};

    \draw[-{Triangle[length=2mm, fill=black, width=1.5mm, line width=0.3mm]}] (2, 0) -- (s2);
    \draw[-{Triangle[length=2mm, fill=black, width=1.5mm, line width=0.3mm]}] (s2) -- (s3);

    \draw[-{Triangle[length=2mm, fill=black, width=1.5mm, line width=0.3mm]}] (s2) -- (a2);

    \draw[-{Triangle[length=2mm, fill=black, width=1.5mm, line width=0.3mm]}] (2, 1) -- (s2);
    \draw[-{Triangle[length=2mm, fill=black, width=1.5mm, line width=0.3mm]}] (a2) -- (s3);

    \draw[-{Triangle[length=2mm, fill=black, width=1.5mm, line width=0.3mm]}] (a2) -- (r2);
    
    \draw[-{Triangle[length=2mm, fill=black, width=1.5mm, line width=0.3mm]}, dashed] (o2) -- (a2);
    \draw[-{Triangle[length=2mm, fill=black, width=1.5mm, line width=0.3mm]}, dashed] (o2) -- (s2);
    \draw[-{Triangle[length=2mm, fill=black, width=1.5mm, line width=0.3mm]}, dashed] (o2) -- (r2);
    
    \draw[-{Triangle[length=2mm, fill=black, width=1.5mm, line width=0.3mm]}] (r2) .. controls (3.75, 0.85) .. (s3);
    
    \draw[-{Triangle[length=2mm, fill=black, width=1.5mm, line width=0.3mm]}] (2, 0.5) -- (s2);
    
    \draw[-{Triangle[length=2mm, fill=black, width=1.5mm, line width=0.3mm]}] (s2) -- ($(r2.south)$);

\end{tikzpicture}}%
        \hspace{-0.5cm}
        \resizebox{0.38\linewidth}{!}{\begin{tikzpicture}

    \draw (2.15, -.75) -- (5.35, -0.75);
    \draw (2.15, -.75) -- (2.15, 3.75);
    \draw (5.35, -.75) -- (5.35, 3.75);
    \draw (2.15, 3.75) -- (5.35, 3.75);
    
    \node[minimum size=1cm] (s1) at (1.45, 0) {};
    \node[circle, draw, double, double distance=0.3mm, minimum size=1cm] (s2) at (3, 0) {$S_t, b_t$};
    \node [minimum size=.5cm] (s3) at (6, -0) {$\dots$};
    
    \node[minimum size=1cm] (a1) at (1.5, 1.5) {};
    \node[circle, draw, minimum size=1cm] (a2) at (4.5, 1.5) {$A_t$};

    \node[] (r1) at (2, 3) {};
    \node[draw, rectangle, minimum size=.8cm] (r2) at (3, 3) {$R_t$};
    
    \node[circle, draw, fill=gray!30, minimum size=1cm] (o2) at (4.5, 3) {$\scriptO_{t:T}$};
    
    \node (tt) at (4.4, -0.55) {$t=0,\dots,T$};

    \draw[-{Triangle[length=2mm, fill=black, width=1.5mm, line width=0.3mm]}] (2, 0) -- (s2);
    \draw[-{Triangle[length=2mm, fill=black, width=1.5mm, line width=0.3mm]}] (s2) -- (s3);

    \draw[-{Triangle[length=2mm, fill=black, width=1.5mm, line width=0.3mm]}] (s2) -- (a2);

    \draw[-{Triangle[length=2mm, fill=black, width=1.5mm, line width=0.3mm]}] (2, 1) -- (s2);
    \draw[-{Triangle[length=2mm, fill=black, width=1.5mm, line width=0.3mm]}] (a2) -- (s3);

    \draw[-{Triangle[length=2mm, fill=black, width=1.5mm, line width=0.3mm]}] (a2) -- (r2);
    
    \draw[-{Triangle[length=2mm, fill=black, width=1.5mm, line width=0.3mm]}] (o2) -- (a2);
    
    \draw[-{Triangle[length=2mm, fill=black, width=1.5mm, line width=0.3mm]}] (r2) .. controls (3.75, 0.85) .. (s3);
    
    \draw[-{Triangle[length=2mm, fill=black, width=1.5mm, line width=0.3mm]}] (2, 0.5) -- (s2);
    
    \draw[-{Triangle[length=2mm, fill=black, width=1.5mm, line width=0.3mm]}] (s2) -- ($(r2.south)$);

\end{tikzpicture}}%
    }
    \caption{
        Generative model (left), and the unstructured (middle) vs. structured inference model (right) that we consider.
        Stochastic variables are placed in circles, deterministic variables in rectangles. 
        The double circle for $\langle S_t, b_t \rangle$ indicates a ``joint'' variable, however, note that $S_t$ is stochastic whereas the belief evolves deterministically. 
        Colored nodes are observed, blank nodes are latent.
    }
    \label{vb:fig:cai_belief}
\end{figure}
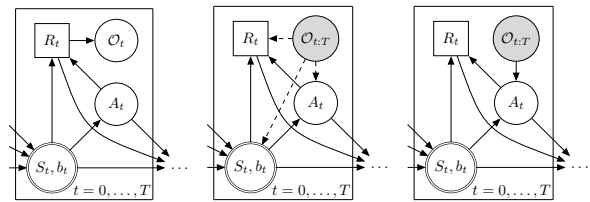

In summary, an EM-style optimization framework can be formulated in our setup by firstly augmenting the generative density $p_\pi(H_{1:T}, M_{1:T})$ (without parameters $\theta$ for brevity, and using the true belief $b_t$) with a binary random variable $\scriptO \in \{0, 1\}$ for each $t$.
We will assume $\scriptO_t \indep H_{1:T} | R_t$, with a choice of likelihood $p(\scriptO_t = 1 | S_t, b_t, A_t) \propto \doubleE_{p(R_t | S_t, b_t, A_t)} \exp R_t$.
Note however, that we will now assume deterministic rewards, $p(R_t | S_t, b_t, A_t) = \delta(R_t - f_R(S_t, b_t, A_t))$, we provide a brief discussion on reward stochasticity in Appendix~\ref{vb:ap:delta_rewards_cai}.
For a finite sequence of data, this corresponds to the generative model in the left of Figure~\ref{vb:fig:cai_belief}.
When we model the inverse problem by conditioning on $\scriptO_{1:T} = 1$, naively we obtain a tilted density,
\begin{align} \label{vb:eq:tilted_history_latents}
    p(H_{1:T}, M_{1:T} | \scriptO_{1:T} = 1) \propto  \Phi_{1:T} \cdot p(H_{1:T}, M_{1:T}),
\end{align}
summarizing the likelihood for $\scriptO_{1:T}$ in terms of a potential $\Phi_{1:T}$.
For brevity, we write $\scriptO_t = 1$ simply as $\scriptO_t$, or $\scriptO_{a:b} = 1$ as $\scriptO_{a:b}$ for sequences.
This inference problem is illustrated in the middle of Figure~\ref{vb:fig:cai_belief}.
Estimating the parameters $\theta$ for this naive model will lead to optimistic learning of the transitions, rewards, and belief \cite{levine2018reinforcement} as the true densities of these components will be shifted towards the potential.
Instead, we consider structuring this inference problem, by focusing on the tilted policy $\pi^+$.

\begin{theorem}[Proof Appendix~\ref{vb:proof:proxy_cai}] \label{vb:theorem:proxy_cai}
    The policy $p_{\pi}^+(A_t | S_t, b_t, \scriptO_{t:T}) \propto \Phi_{t:T} \cdot \pi^+(A_t | S_t, b_t)$ satisfies,
    \begin{align}
        q^* &= \argmax_q \scriptJ_{\textit{proxy}}(q, \pi^+) \nonumber\\
        &= \argmax_q \doubleE_{p_q}\left[\sum_{t=1}^T R_t - \kldiv(q(A_t | S_t, b_t) \Vert \pi^+) \right] \nonumber
    \end{align}
     which is a regularized policy objective, where $q^*$ also guarantees a policy improvement in the unregularized MDP, $\doubleE_{p_{q^*}} [\sum_{t=1}^T R_t] \ge \doubleE_{p_{\pi^+}} [\sum_{t=1}^{T} R_t].$
\end{theorem}

The inference problem posed by Theorem~\ref{vb:theorem:proxy_cai} belongs to the graphical model on the right of Figure~\ref{vb:fig:cai_belief}, and gives us a proxy objective $\scriptJ_{\textit{proxy}}(q, \pi^+)$ for the RL-problem that regularizes $q^*$ to $\pi^+$ in KL-divergence.
We can use this to naturally formulate an approximate policy iteration algorithm through the EM-framework, as given in Proposition~\ref{vb:prop:em}.

\begin{proposition}[Proof App.~\ref{vb:proof:em}] \label{vb:prop:em}
    Starting from some prior $\pi^+_{(1)}$ with sufficient density everywhere, at each iteration $i$, alternate the following: \textit{E-step}: 
    solve $\max_q \scriptJ_{\textit{proxy}}(q, \pi^+_{(i)})$ to get $q^*_{(i+1)}$ by fixing the current prior $\pi^+_{(i)}$; 
    \textit{M-step}: distill the solution $q^*_{(i+1)}$ into a new prior $\pi^+_{(i+1)}$ by solving $\max_{\pi^+} \scriptJ_{\textit{proxy}}(q^*_{(i+1)}, \pi^+)$.
    The sequence $\{\pi_{(i)}^+\}_{i=1}^N$ converges to the Bayes-optimal policy.
\end{proposition}

Actually solving each alternating step for the EM-procedure of Proposition~\ref{vb:prop:em} is usually intractable.
To mitigate this, the M-step can be approximated by parameterizing the prior policy with parameters $\pi^+_\theta$ and performing a few steps of gradient ascent on $\nabla_\theta \scriptJ_{\textit{proxy}}(q^*_{(i+1)}, \pi^+_\theta)$ \cite{bishop2007}.
In MDP-algorithms like MPO \cite{abdolmaleki2018maximum}, the E-step is approximated by estimating the objective $\scriptJ(q, \pi)$ through learning of a value function $Q^{q, \pi}_{\textit{soft}}(S_t, A_t)$ from sample rollouts. 
The `soft' subscript emphasizes that this value function estimates the $q$ to $\pi$ KL-regularized value.
Then, a 1-step optimization problem is performed to obtain sample solutions to the tilted policy $\hat{q}^* \approx \pi \exp Q^{q, \pi}_{\textit{soft}}$ analogous to Theorem~\ref{vb:theorem:proxy_cai} over some stationary distribution of states. 

\section{Bayes-Adaptive SMC-Planning}

We present an Expectation-Maximization (EM) framework to learn solutions to Bayes-adaptive MDPs through a combination of variational Bayes, sequential Monte-Carlo planning, and meta-learning.
On a base level, we make a considerable improvement to the E-step by addressing a bottleneck induced by gradient-based learning of variational belief-states.
Instead of directly sampling from our policy as done in model-free methods, which is cheap, we shift more of the computation to online planning to generate higher quality data and learning targets.
This improves scaling to compute resources for learning adaptive agents compared to model-free approaches, because we assume that the bottleneck of training is induced by stochastic gradient descent for which planning can enhance this procedure at reasonable cost.
This is a reasonable assumptions when training with long histories due to the $O(1)$ inference cost of state-space sequence models at decision-events, but at least $O(\log n)$ costs during BPTT.
To achieve this, we adapt the Control-as-Inference framework \cite{ziebart_modeling_2010, levine2018reinforcement} to our setting to derive a new sequential importance resampling procedure \cite{piche2018probabilistic} in Section~\ref{vb:sec:belief_smc} and an EM-formulation for belief learning in Section~\ref{vb:sec:varbelief_learning}.

\subsection{E-step: Belief-SMC planning} \label{vb:sec:belief_smc}

In MDP-settings, typical methods \cite{abdolmaleki2018maximum, macfarlane2024spo} make strong use of the duality from Theorem~\ref{vb:theorem:proxy_cai} to obtain approximate solutions, it implies that the optimal regularized policy requires us to infer the tilted policy $p^+_\pi(A_t | S_t, b_t, \scriptO_{t:T}), \forall s \in \scriptS$.
We will also use this duality to sample solutions $\hat{q}_t^* \approx q^*_t = p_\pi^+(A_t | S_t, b_t, \scriptO_{t:T})$ through a particle-filter planner with variational beliefs, which we can subsequently use as learning targets in the M-step.
As stated in the Introduction, we want to perform importance-sampling to an intractable \emph{target} distribution with a more manageable \emph{proposal} distribution, and particle-filter methods provide tools on how to efficiently achieve this \cite{chopin_introduction_2020}.

Suppose we are interested in the cumulative returns of an agent over a window, $f(H_{1:t}) = \sum_{i=1}^t R_i$, under the target distribution given in \myeqref{vb:eq:tilted_history_latents}.
We can then write importance-sampling weights as,
\begin{align}
    \doubleE_{\Phi_{1:T} p_\pi} f(H_{1:t}) 
    &= 
    \doubleE_{p_{q, b^q}(H_{1:T}, M_{1:T})} [w_t \cdot f(H_{1:t})]
    \\
    w_t &= \frac{p_\pi(H_{1:T}, M_{1:T}| \scriptO_{1:T})}{p_{q, b^q}(H_{1:T}, M_{1:T})}
\end{align}
with proposal $p_{q, b^q}(H_{1:T}, M_{1:T})$ being similar to \myeqref{vb:eq:rl_history_latents},
\begin{align} \label{vb:eq:proposal}
    p_{q, b^q}(H_{1:T}, M_{1:T}) &= \doubleE_{\rho} \Big[ \prod_{t=1}^T p(S_{t+1}, R_t | A_{t}, S_t, M_t) 
    \nonumber \\
    &\cdot\:q(A_t | S_t, M_t) \cdot q(M_t | H_{<t}) \Big],
\end{align}
where we assume knowledge of $\rho(S_1, M_0)$.
We call $b_t^q(M) = q(M_t | H_{<t})$ the variational belief with $b_t$ the true belief, and $q(A_t | S_t, M_t)$ the variational policy with $\pi^+$ the prior policy.
The idea is that we sample sequences according to our variational model $q$, and weight the resulting statistics according to their importance weights $w_t$.

\begin{corollary}[Proof Appendix~\ref{vb:proof:is_weights}] \label{vb:cor:is_weights}
    The importance sampling weights $w_t$ for drawing approximate samples from \myeqref{vb:eq:tilted_history_latents} using the proposal distribution from \myeqref{vb:eq:proposal} can be expressed as,
    \begin{align*}
        \frac{w_t}{w_{t-1}} = \doubleE_{b^q_t}
        \doubleE_{p_q} \biggr[
            \frac{\doubleE_{b_t^q} \omega_t p_{\pi}(H_t | S_t, M_t)}{\doubleE_{b^q_t} p_q(H_t | S_t, M_t)}
            \cdot \exp R_t
            \nonumber \\
            \cdot \:\frac{
                \doubleE_{b_{t+1}^q} \omega_{t+1} \exp V_{\mathit{soft}}^{q, \pi}(S_{t+1}, M_{t+1})
            }{\doubleE_{b_t^q} \omega_t \exp V_{\mathit{soft}}^{q, \pi}(S_t, M_t)}
            \cdot f(H_{1:t})   
        \biggr]
    \end{align*}
    where $\omega_t = \frac{b_t(M)}{b_t^q(M)}$ are nested importance-sampling weights to adjust for the variational beliefs.
\end{corollary}

There are two main limitations to evaluating the weights in Corollary~\ref{vb:cor:is_weights}, we need the regularized value functions $V^{q, \pi}_{\mathit{soft}}$ and the true belief $b_t$ to evaluate the nested weights $\omega_t$.
To deal with this, we will learn the value functions $V_\theta$ using Temporal-Difference methods \cite{sutton_reinforcement_2018, lawson2018twisted} as is typical in SMC-planners \cite{piche2018probabilistic, lioutas2023critic, macfarlane2024spo, vries2025trustregion}.
Then, we will approximate the nested weights $\hat{\omega}_t \approx \omega_t$ with a single-timestep correction derived from the belief state recursion,
\begin{align} \label{vb:eq:belief_correction}
    b_{t+1}(M) \propto b_t(M) \cdot p(H_t | M) \approx b^q_t(M)\cdot p(H_t | M)
\end{align}
In other words, we will treat the previous variational belief $b_t^q$ as if it were the true belief at the previous timestep. 
This is also a common heuristic employed in variational sequence models to regularize their model's predictions in a way that remains computationally feasible \cite{hafner2020dream}.
From the belief recursion, we reuse importance sampling tools to write,
\begin{align}
    \doubleE_{b_t^q} \frac{b_t(M_t)}{b_t^q(M_t)} e^{V_{\mathit{soft}}^\pi(S_t, M_t)} \approx \sum_{j=1}^{K_\Omega} \bar{\omega}_t^{(j)} e^{V_{\mathit{soft}}^\pi(S_t, M_t^{(j)})}, 
\end{align}
where $M^{(j)}_t \sim b^q_t$, and $\ln \hat{\omega}_t^{(j)}$ are the belief-particles, such that $\bar{\omega}_t^{(j)} = \text{softmax}(\hat{\omega}_t)^{(j)}$ normalizes,
\begin{align}
    \ln \hat{\omega}_t^{(j)} &= \ln \frac{\hat{b}_{t-1}^q(M_t^{(j)})}{ b_t^q(M_t^{(j)})} + \ln p(H_t | S_t, M_t^{(j)})
\end{align}
Finally, we correlate (re-use) samples $M \sim b_t^q$ for each term involving $\omega_t$ in Corollary~\ref{vb:cor:is_weights} to reduce variance between independent model components.

Our approach differs from POMDP methods in that we regenerate all nested particles $\omega_t^{(j)}$ at each timestep from the variational beliefs, rather than let them evolve them through a transition function \cite{abdulsamad2025sequentialmontecarlopolicy}.
A full overview of our variational Bayes-adaptive planner is given in pseudocode in Algorithm~\ref{vb:alg:ba_particle_filter}.
The pseudocode makes use of variational paramerization of the beliefs $b_{\phi_t}^q(M)$, with an inference function $h_\theta$ to generate $\phi_t$, this part and the M-step (learning) are explained in the next section.


\begin{figure}[t]
    \centering
    \begin{minipage}[t]{0.49\textwidth}
        \input{varibased/figures/ba_smc_pseudocode}
    \end{minipage}
\end{figure}

\subsection{Gradient Based (EM)-step for Amortized Variational Bayes}  \label{vb:sec:varbelief_learning}

The previous section described how we can sample approximate solutions $\hat{q}^*$ to the E-step for the policy $\pi_\theta$.
The EM-loop is then completed for $\pi_\theta$ by minimizing the empirical cross-entropy $\min_\theta \doubleE[\mathit{CE}(\hat{q}^*, \pi_\theta)]$.
However, our approach also relies on variational belief states $b^q_t$, which induces a joint optimization problem (i.e., for $b^q$ and $\pi_\theta$).
Firstly, rather than estimating $b^q$ at each timestep explicitly, we assume a simple parametric form with parameters $\phi_t$, such that $b^q_t \leftarrow b^q_{\phi_t}$ where $\phi_t$ can be generated using an autoregressive inference function $\phi_t = h_\theta(\phi_{t-1}, H_{t-1})$.
For instance, when $b^q_{\phi_t}$ is assumed to be Gaussian, we can use a recurrent neural network \cite{hochreiter_long_1997} or state-space model \cite{lu_s5rl_2023} for $h_\theta$ to predict the mean and covariance $\phi_t = (\mu_t, \Sigma_t)$ of this latent distribution.

Although approximating the E-step through sampling makes sense for $\pi_\theta$ due to the non-differentiable operations in the particle-filter planner (e.g., resampling in Algorithm~\ref{vb:alg:ba_particle_filter}), this is not necessarily the best strategy for the beliefs $b^q_{\phi_t}$ \cite{gu_neural_2015}.
To reduce variance, we will instead opt for an amortized gradient-based optimization of the E-step for obtaining $\phi_t^*$ \cite{kingma2022autoencodingvariationalbayes}.
This requires us to formulate a training objective through the evidence lower-bound in order to optimize for $\theta^*_{(n)}$, which conveniently reduces the M-step to a simple assignment $\theta_{(n+1)} \leftarrow \theta_{(n)}^*$.
To formulate this E-step, we first state the following proposition, which is essentially an extension and rearrangement of Theorem~\ref{vb:theorem:proxy_cai}.



\begin{proposition}[Proof Appendix~\ref{vb:proof:kl_to_elbo_swap}] \label{vb:prop:kl_to_elbo_swap}
    The optimization objective $\max_{b^q_t, q} \scriptJ((b_t^q, q_t), (b_t, \pi^+_t))$ with now the variational and true belief respectively, can be written as,
    \begin{align*}
        \doubleE_{\rho_{q^\star}} 
        \max_{b_t^q} 
        \doubleE_{
            b_t^{q}
        }
        (
            \max_{q_t} 
            \doubleE_{
                q_t
            }
            [
                &Q^{q^\star}(S_t, b_t, A_t)
                -
                \kldiv(q_t \Vert \pi^+_t)
            ]
            \\
            & +
            \scriptL_\scriptZ(b_t^q | b_0)
        ),
    \end{align*}
    where $\scriptL_\scriptZ(b_t^q | b_0)$ is another evidence lower-bound for the marginal $p(H_{<t})$, defined as,
    \begin{align*}
        \scriptL_\scriptZ(b^q_{t+1} | b_0) = 
        -\kldiv(b^q_{t+1} \Vert b_0) 
        + 
        \doubleE_{
            b^q_{t+1}
        }
        \biggr[
            \ln p_\theta(S_1 | M)
            \\
            +\sum_{i=1}^t
            \ln p_\theta(S_{t+1}, R_t | S_t, A_t, M) 
            +
            \ln \pi_\theta(A_t | S_t, M)
        \biggr],\nonumber
    \end{align*}
    where we assume $b_0 = p(M_0)$ to be a simple fixed prior.
\end{proposition}

The above result is an important property as it shows the relation of policy search $q_t$ to the variational belief search $b^q_t$.
It shows that we can sample trajectories $H_{<t} \sim \rho_{q^*}$ from the optimal history-occupancy to then a) update our belief $b^q_t$ and do planning to infer $q^*_t$ (Alg.~\ref{vb:alg:ba_particle_filter}), and b) perform gradient learning to amortize the nested maximization.

From Prop.~\ref{vb:prop:kl_to_elbo_swap}, we can motivate the following loss function to fit our joint neural net parameters $\theta$ to the graphical model depicted on the right of Figure~\ref{vb:fig:cai_belief},
\begin{align}
    \scriptL(\theta; \scriptD_{(n)}) 
    = 
    \doubleE_{\hat{V}_t, \hat{q}^*_t, H_{t-h:t} \sim \scriptD_{(n)}}[
        \frac{c_v}{2} (\hat{V}_t - V_\theta^\pi(S_t, \textbf{sg}[\phi_t]))^2 
    \nonumber \\
        - c_\pi \doubleE_{A_t \sim \hat{q}^*_t}[\ln \pi_\theta^+(A_t | S_t, \textbf{sg}[\phi_t]))]
        -
        c_b\scriptL_\scriptZ(\phi_t | \phi'_{t-h})
    ]\nonumber
\end{align}
where $\phi_t = h_\theta(\phi_{t-h}; H_{t-h:t})$ is the unrolled variational parameters over a burn-in window, $\phi_{t-h}'$ is a cached hidden state, and $\textbf{sg}[\cdot]$ is a stop-gradient operator for the reparametrized predictive for the value and policy networks.
In other words, we have replaced the base prior $b_0$ from Prop.~\ref{vb:prop:kl_to_elbo_swap} with an intermediate prior `$h$' steps before the current timestep $t$.
This last unrolling and hidden state caching is important as it efficiently deals with varying length trajectories when optimizing this loss on the GPU, which requires fixed-size batches.
However, the recurrent state caching can drift in representation from the optimization process as noticed by \citet{Kirkpatrick_2017}.
In practice, this typically requires longer unroll windows to deal with this mismatch, which is a trade off between performance and runtime.

Our final heuristic we use in optimizing our loss function is to cancel the gradients for $\pi_\theta$ in the belief ELBO $\scriptL_\scriptZ(\phi_t | \phi'_{t-h})$.
This means that the variational belief is regularized to the term $\ln \pi_{\text{sg}[\theta]}(A_t | S_t, M)$ in a way that does not influence the policy network (see Prop.~\ref{vb:prop:kl_to_elbo_swap}).
This is motivated by the fact that we are already learning $\pi_\theta$ on the sample policies from the SMC planner; we want to prevent a double learning effect while also regularizing the belief to support the policy predictions.
The full (outer-) learning loop is provided in Algorithm~\ref{alg:pseudocodeouter} in the Appendix.

\section{Related Work}
For brevity, Appendix~\ref{app:related_work} extends our related work.

\paragraph{Belief Learning in RL.}
Our approach builds on sequence-based approximate belief learning \cite{igl_deepvar_2018, becker2024kalmambaefficientprobabilisticstate}, specifically leveraging the S5 state-space model \cite{lu_s5rl_2023, smith2023simplifiedstatespacelayers} for its efficiency in long-horizon problems.
Most closely related is VariBAD \cite{zintgraf_varibad_2020}, which separately optimizes a belief model and a policy via an ELBO objective. 
In contrast, we treat the policy as an intrinsic part of the probabilistic graphical model, resulting in a natural end-to-end regularized learning algorithm.
While recent extensions to VariBAD have incorporated online planning \cite{jeong2025reflectthenplan}, they did not evolve the posterior during planning. 

\paragraph{SMC Planning.}
Sequential Monte Carlo (SMC) has emerged as a powerful tool for inferring regularized policies in RL \cite{macfarlane2024spo} and LLM steering \cite{lew2023sequentialmontecarlosteering}. While early Variational SMC methods required complex differentiation through non-differentiable resampling \cite{naesseth2018varSMC}, recent work avoids this by using Fisher's identity for policy optimization \cite{abdulsamad2025sequentialmontecarlopolicy}. We adopt an EM-style framework \cite{gu_neural_2015, macfarlane2024spo} where SMC acts as a probabilistic planner \cite{piche2018probabilistic}. Unlike prior planners that assume perfect models \cite{vries2025trustregion}, our method explicitly addresses the model mismatch between the learned variational belief and generative dynamics by adapting the importance sampling weights.

\section{Experiments}

We present results comparing our method, VariBASeD, against the model-free RL$^2$ baseline \cite{duan_rl2_2016} on two environments: function optimization and gridworld exploration.
We consider a multi-task setting with a uniform, but unknown, prior distribution over tasks. 
The specific details of this task distribution are described in Appendix~\ref{vb:ap:setup:envs} along with hardware requirements in Appendix~\ref{vb:ap:hardware}, and algorithm hyperparameters in Appendix~\ref{vb:ap:hyper}. 
Our experiments aim to demonstrate: a) the ability of VariBASeD to achieve effective (in-domain) test-time adaptation, and b) that our method's performance scales with increased planning budget relative to both data and runtime complexity.

\begin{figure*}[ht]
    \centering
    \begin{minipage}{0.45\linewidth}
        \centering
        \includegraphics[width=\linewidth]{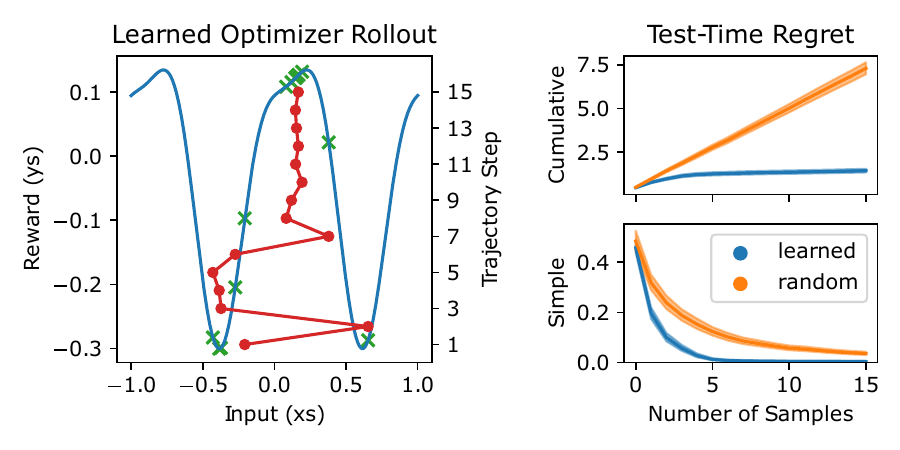}
    \end{minipage}%
    \hspace{\fill}
    \begin{minipage}{0.45\linewidth}
        \centering
        \includegraphics[width=\linewidth]{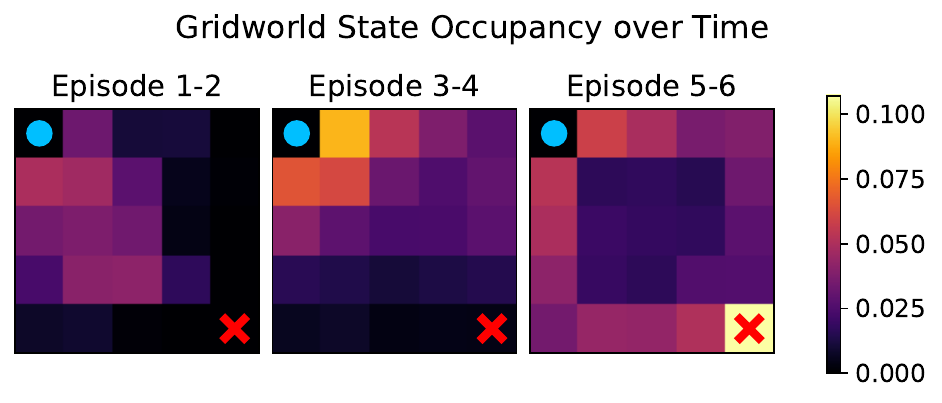}
    \end{minipage}
    \caption{Evaluation rollouts on the function optimization and gridworld problems over multiple test-episodes. 
    Due to the stochastic nature of our method, we visualize one sample rollout for the function optimization problem (with average metrics), and the averaged rollouts for the gridworld. 
    For the gridworld, the circle indicates the starting-tile and the cross the goal-tile.
    On the function optimization we used a planning budget of $H=1, K=16$ and on the gridworld $H=4, K=32$.
    }
    \label{vb:fig:didactic_test}
\end{figure*}

\paragraph{Example Rollouts.} 
In Figure~\ref{vb:fig:didactic_test}, we first show example rollouts of a single fully trained agent on the two considered problems.
Due to the stochastic nature of our particle planner, we provide a visualization of test rollouts for both a single sample rollout and the full distribution's state-occupancy.
Specifically, for the function optimization problem, we show a single agent rollout (red line) whereas for the gridworld problem we show its rollout distribution in terms of the state-occupancy.
Compared to a random baseline, our method demonstrates adaptive behavior on the test task in both environments.
This is shown by the state-occupancy on the gridworld concentrating on the goal tile in episodes 5–6, and the sublinear cumulative regret achieved on average in the function optimization problem.

\begin{figure*}
    \centering
    \includegraphics[width=0.49\linewidth]{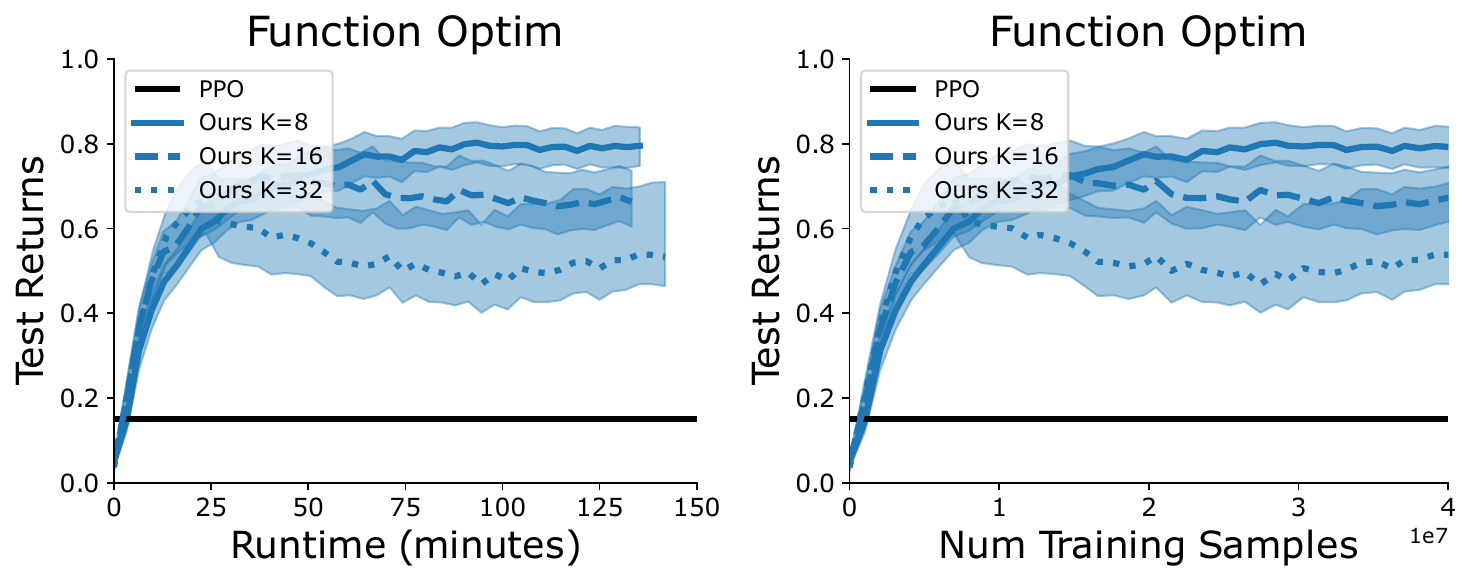}
    \hspace{\fill}
    \includegraphics[width=0.49\linewidth]{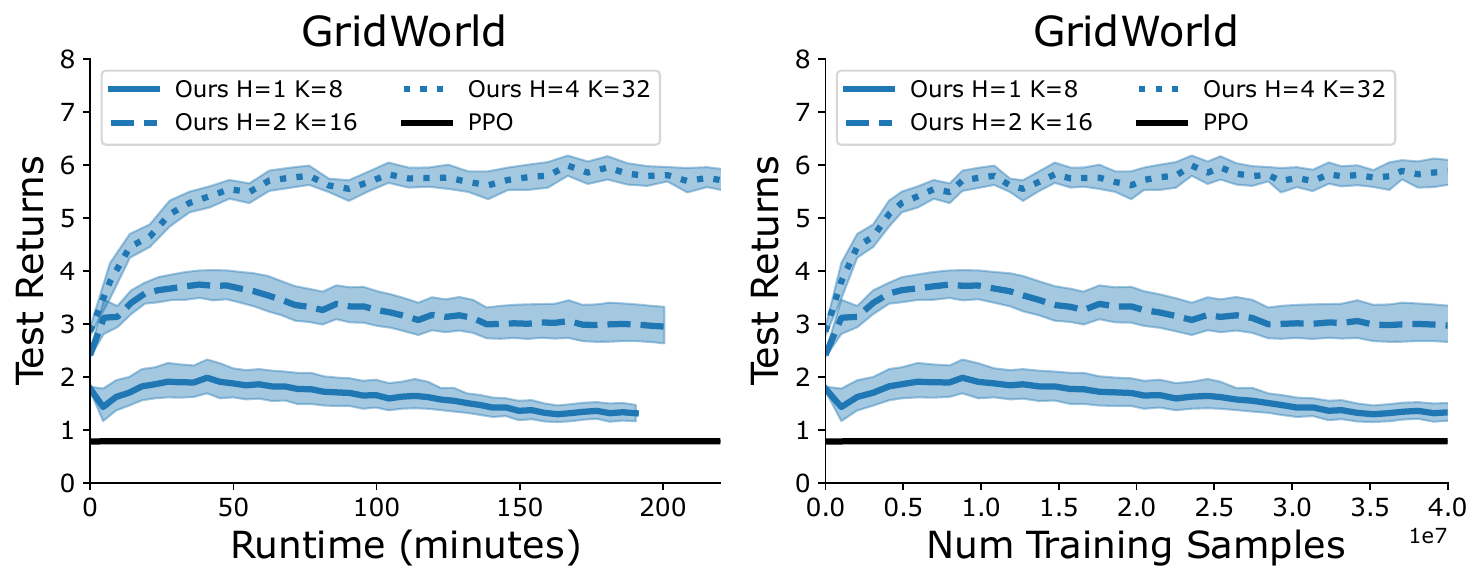}
    \caption{Learning curves for all environments comparing our VariBASeD against recurrent PPO (RL$^2$) \cite{duan_rl2_2016} using the S5 architecture \cite{lu_s5rl_2023} for different planning budgets.
    Shaded regions give 99\% two-sided BCa-bootstrap intervals over 30 seeds.
    In our framework we found that the PPO baseline did not learn.}
    \label{vb:fig:main_results}
\end{figure*}

\paragraph{Learning Curves.} 
Whereas Figure~\ref{vb:fig:didactic_test} shows results for single trained agents, we also plot the learning curves over multiple training repetitions in Figure~\ref{vb:fig:main_results}.
Specifically we show how the learning curves are affected by planning budget.
On the function optimization problem we used a search-depth of $H=1$, with a particle budget of $K \in \{8, 16, 32\}$.
In contrast, on the gridworld, we used a coupled planning budget of $\langle H=1,K=8 \rangle$, $\langle H=2,K=16\rangle$, and $\langle H=4,K=32\rangle$.
We included the PPO (RL$^2$) baseline for reference; however, it did not learn effectively in our setup.

Interestingly, we find an opposing pattern on each environment when comparing the effect of the planner budget.
As hypothesized, on the gridworld problem, we find that increasing planner budget significantly improves the agent's efficacy.
In contrast, increasing planner budget slightly reduced agent performance on the function optimization problem, which goes against our hypothesis.
Furthermore, we also find that all learning curves quickly level-off and begin to slightly deteriorate as learning continues.
Although this finding supports our second claim, the impact is limited since lower budgets do not reach optimality.

\paragraph{Discussion.} 
We hypothesize that our VariBASeD agent does not enjoy much from increased planner budget on the function optimization problem due to an additional exploration effect introduced by using fewer particles.
The smaller particle budget adds noise into the overall learning process, which induces more diverse data collection and prevents premature convergence to a suboptimal policy.

This pattern does not hold on the gridworld problem which strongly benefits from additional budget.
In contrast to the function optimization, the gridworld environment has actual state-dynamics, and additional lookahead enables the planner to make better use of the value and reward models.
We expect that the gridworld requires additional budget for accurate policy inference, whereas for function optimization, a lower budget was sufficient, and the increased budget likely resulted in overly greedy action selection.

Overall, we suspect that the complex loss landscape and caching of the recurrent neural network state causes a representational misalignment that impedes effective learning with our current architecture.
This hypothesis implies that simulation based inference is crucial for enabling good performance in our framework. 
This routine corrects the mispredictions from the neural networks through a proper inference step while still leveraging the efficiency benefits of having an approximate guide during search.
If this is true, that would explain the poor result obtained by our PPO agent, since the loss landscape could be too difficult for gradient descent to navigate for obtaining a proper policy improvement.
At the same time, PPO's performance could also be attributed to poor tuning, which resulted in a too difficult optimization landscape. 
With that said, our approach shows a simpler yet effective way to obtain policy improvement by simply scaling up the planner budget.

\section{Conclusion}
We presented VariBASeD, a scalable and holistic framework for approximate Bayes-adaptive planning that combines sequential Monte-Carlo inference with amortized variational learning of belief states. 
By coupling online SMC planning with gradient-based belief optimization, our method is posed to improve both sample and runtime efficiency over model-free baselines.
We hypothesized that this is due to extracting more informative learning target per inference step, which ends up accelerating the entire training pipeline.
Preliminary experiments indeed demonstrate that shifting computational effort from training to online planning yields improved performance in small scale meta-reinforcement learning settings. 
Future work will test domains with longer context lengths and higher task variety to demonstrate the potency of our method on large-scale adaptive control.

\section*{Acknowledgements}
JdV, MdW, and MS are supported by the AI4b.io program, a collaboration between TU Delft and dsm-firmenich, which is fully funded by dsm-firmenich and the RVO (Rijksdienst voor Ondernemend Nederland).

\printbibliography

\newpage
\onecolumn

\title{Supplementary Material}
\maketitle

\appendix
\section*{Appendix}

\section{Extended Related Work} \label{app:related_work}

\paragraph{State-Space Models for RL.}
The choice of S5 \cite{smith2023simplifiedstatespacelayers} as our backbone is motivated by the limitations of traditional architectures. While RNNs struggle with long-range dependencies, and Transformers \cite{parisotto2021efficient} face quadratic complexity with history length, SSMs utilize the associativity of hidden state dynamics to enable training-time parallelism. Furthermore, S5 allows for efficient inference by storing compact hidden states between batches, facilitating the truncation of histories without the significant performance penalties associated with Transformer-based caches.

\paragraph{Variational Inference and Policy Control.}
The intersection of Variational Inference (VI) and control has been studied in the RL as inference framework \cite{levine2018reinforcement, ziebart_maxent_2010}, where the optimal policy is the one that minimizes the KL-divergence to a target distribution. This target distribution amounts to a KL-regularized optimal policy in the classical RL sense \cite{sutton_reinforcement_2018}. \citet{abdulsamad2025sequentialmontecarlopolicy} investigate a similar target distribution to ours, for which they utilize nested sequential Monte-Carlo to approximate a tilted variational belief distribution. 
Instead, we use SMC planning to perform online policy improvement.

\paragraph{Applications of SMC in Diverse Domains.}
Beyond traditional robotics, SMC-based planning is gaining traction in Bayesian experimental design \cite{iqbal2024nestingparticlefiltersexperimental} to optimize information gain. Our derivation of importance sampling (IS) weights attempts to remove complicated nested expectations typically required for value function estimation in stochastic transitions.

\section{Derivations \& Proofs} \label{vb:ap:proofs}

\begin{numproof}
    (Theorem~\ref{vb:theorem:proxy_cai}) \label{vb:proof:proxy_cai}
    We want to confirm whether $q^* = \argmax_q \scriptJ_{\mathit{proxy}}(q, \pi^+)$ is consistent with $p_{\pi}^+(A_t | S_t, b_t, \scriptO_{t:T})$.
    If we write this in terms of the full trajectory distribution, we get,
    \begin{align}
        \min_{q} \kldiv (p_q(H_t) \Vert p_{\pi}^+(H_t | \scriptO_{t:T})).
    \end{align}
    Notably, we have extended the states $S_t$ with the belief state $b_t$ to obtain an augmented state-space $\langle S_t, b_t \rangle \in \scriptS \times \scriptB$.
    If we then assume that the augmented dynamics and rewards follow the definition stated in the background,
    \begin{align}
        p_{S}^+(b_{t+1}, S_{t+1} | S_t, A_t, b_t) = \delta(b_{t+1} = p(M_{t+1} | H_{<t} \cup H_t))\doubleE_{M_t \sim b_t}p(S_{t+1} | A_t, S_t, M_t), \nonumber
        \\[6pt] 
        p_{R}^+(R_{t} | S_t, A_t, b_t) = \doubleE_{M_t \sim b_t}p(R_t | A_t, S_t, M_t), \quad
        \pi^+(A_{t} | S_t, b_t) = \doubleE_{M_t \sim b_t}\pi(A_t | S_t, M_t). \nonumber
    \end{align}
    Then the stated Theorem reduces to a reformulation of Theorem~2.2 from \citet{vries2025trustregion} with a modified reward and transition density in the graphical model. 
    Thus, the proof follows identically and we can safely state that $q^*(A_t | S_t, b_t) = p^+_\pi(A_t | S_t, b_t, \scriptO_{t:T})$.
\end{numproof}

\begin{numproof}
    (Proposition~\ref{vb:prop:em}) \label{vb:proof:em}
    Similar to Theorem~\ref{vb:theorem:proxy_cai}, this is a reformulation of the dynamics and reward function of the probabilistic graphical model discussed in \cite{vries2025trustregion} and the proof directly follows from typical Expectation-Maximization derivations \cite{Neal1998}.
\end{numproof}

\begin{numproof}
    (Corollary~\ref{vb:cor:is_weights}) \label{vb:proof:is_weights}
    Using the target definition from \myeqref{vb:eq:tilted_history_latents} and the imposed structure by Theorem~\ref{vb:theorem:proxy_cai}, we first manipulate the marginal density $p_\pi(H_{1:t})$ to obtain a weight-recursion in terms of history dependence, 
    \begin{align}
        \doubleE_{p_\pi(H_{1:t} | \scriptO_{1:T})} f(H_{1:t}) 
        &= 
        \doubleE_{p_q(H_{1:t})} 
        \frac{p_\pi(H_{1:t} | \scriptO_{1:T})}{p_q(H_{1:t})} f(H_{1:t})
        = 
        \doubleE_{p_q(H_{1:t})} 
        [w_t \cdot f(H_{1:t})]
        \\
        &=
        \doubleE_{p_q(H_{1:t})} \left[
            \frac{p_\pi(H_{t} | H_{1:t-1}, \scriptO_{t:T}) p_\pi(H_{1:t-1} | \scriptO_{1:T})}{p_q(H_{t | H_{1:t-1}})p_q(H_{1:t-1})} f(H_{1:t})    
        \right]
        \\
        &=
        \doubleE_{p_q(H_{1:t})} \left[
            w_{t-1} 
            \frac{p_\pi(H_{t} | H_{1:t-1}, \scriptO_{t:T})}{p_q(H_{t | H_{1:t-1}})} 
            f(H_{1:t})    
        \right]
        \\
        &=
        \doubleE_{p_q(H_{1:t-1})} 
        \left(
            w_{t-1} 
            \doubleE_{p_q(H_t | H_{1:t-1})} \left[
                \frac{p_{\pi}(H_t | H_{1:t-1}, \scriptO_{t:T})}{p_q(H_t | H_{1:t-1})} 
                f(H_{1:t})    
            \right]
        \right)
    \end{align}
    Now extract the $w_t$ term for brevity and separate the conditioning on $\scriptO$,
    \begin{align}
        w_t &= w_{t-1} \cdot
        \doubleE_{p_q(H_{t}|H_{1:t-1})} \left[
            \frac{p_{\pi}(S_{t+1}, R_t, A_t | H_{1:t-1}, \scriptO_{t:T})}{p_q(S_{t+1}, R_t, A_t | H_{1:t-1})} 
            \cdot 
            f(H_{1:t})   
        \right]
        \\
        &=w_{t-1} \cdot
        \doubleE_{p_q} \left[
            \frac{p_{\pi}(S_{t+1}, R_t, A_t | H_{1:t-1})}{p_q(S_{t+1}, R_t, A_t | H_{1:t-1})} 
            \cdot 
            \frac{p_{\pi}(\scriptO_{t:T} | S_{t+1}, R_t, A_t, H_{1:t-1})}{p_\pi(\scriptO_{t:T} | H_{1:t-1})}
            \cdot 
            f(H_{1:t})   
        \right]
        \\
        &=w_{t-1} \cdot
        \doubleE_{p_q} \left[
            \frac{p_{\pi}(S_{t+1}, R_t, A_t | H_{1:t-1})}{p_q(S_{t+1}, R_t, A_t | H_{1:t-1})} 
            \cdot 
            \frac{p(\scriptO_{t} | R_t)p_{\pi}(\scriptO_{t+1:T} | H_{1:t})}{p_\pi(\scriptO_{t:T} | H_{1:t-1})}
            \cdot 
            f(H_{1:t})   
        \right]
        \\
        &\propto w_{t-1} \cdot
        \doubleE_{p_q} \left[
            \frac{p_{\pi}(S_{t+1}, R_t, A_t | H_{1:t-1})}{p_q(S_{t+1}, R_t, A_t | H_{1:t-1})} 
            \cdot 
            \exp R_t \cdot \frac{\exp V^{q, \pi}_{\mathit{soft}}(H_{1:t})}{\exp V^{q, \pi}_{\mathit{soft}}(H_{1:t-1})}
            \cdot 
            f(H_{1:t})   
        \right]
    \end{align}
    where the proportionality is due to $p(\scriptO_t | R_t) \propto \exp R_t$ and where $V^{q, \pi}_{\mathit{soft}}(H_{1:t}) = \ln p_{\pi}(\scriptO_{t+1:T} | H_{1:t})$. 
    For brevity, assume that $p(\scriptO_t | R_t) = \exp R_t$ properly normalizes as is from this point on.

    Now to deal with history-dependence, we will insert the true and variational belief states,
    \begin{align}
        \frac{w_t}{w_{t-1}} 
        =
        \doubleE_{p_q(H_t | S_t, b_t^q)} \left[
            \frac{p_{\pi}(S_{t+1}, R_t, A_t | S_t, b_t)}{p_q(S_{t+1}, R_t, A_t | S_t, b_t^q)} 
            \cdot 
            \exp R_t \cdot \frac{ \exp V^{q, \pi}_{\mathit{soft}}(S_{t+1}, b_{t+1})}{\exp V^{q, \pi}_{\mathit{soft}}(S_t, b_t)}
            \cdot 
            f(H_{1:t}) 
        \right].
    \end{align}
    Observe that $\doubleE_{p_q(H_t | S_t, b_t^q)}(\cdot) = \doubleE_{\doubleE_{M \sim b_t^q} p_q(H_t | S_t, M)}(\cdot)$ per definition (see Background), this can be rewritten more simply as $\doubleE_{b_t^q(M)}\doubleE_{p_q(H_t | S_t, b_t^q)}(\cdot)$ using Fubini's theorem.

    Reparametrize the value functions $V^{q, \pi}_{\mathit{soft}}(H_{1:t-1}) = V^{q, \pi}_{\mathit{soft}}(S_t, b_t) = \doubleE_{M \sim b_t} V^{q, \pi}_{\mathit{soft}}(S_t, M)$ to obtain,
    \begin{align}
        \frac{w_t}{w_{t-1}} 
        =
        \doubleE_{b_t^q}\doubleE_{p_q} \left[
            \frac{\doubleE_{b_t}p_{\pi}(S_{t+1}, R_t, A_t | S_t, M_t)}{\doubleE_{b_t^q}p_q(S_{t+1}, R_t, A_t | S_t, M_t)} 
            \exp R_t 
            \frac{\doubleE_{b_{t+1}}\exp V^{q, \pi}_{\mathit{soft}}(S_{t+1}, M_{t+1})}{\doubleE_{b_t}\exp V^{q, \pi}_{\mathit{soft}}(S_t, M_t)}
            f(H_{1:t}) 
        \right].
    \end{align}
    Importance sampling with $b_t^q$, s.t., $\omega_t = \frac{b_t(M)}{b_t^q(M)}$ for all terms involving $\doubleE_{b_t}(\cdot)$ obtains the result.
\end{numproof}

\begin{numproof}(Proposition~\ref{vb:prop:kl_to_elbo_swap}) \label{vb:proof:kl_to_elbo_swap}
    The first part of the stated lower-bound (optimization objective) restates the property of Proposition~\ref{vb:theorem:proxy_cai}, which only leaves the right-term.
    The KL-divergence has a well known decomposition into an lower-bound and evidence term,
    \begin{align}
        \scriptL_\scriptZ(b^q_{t+1}) 
        = \kldiv(b^q_t \Vert b_t) &=
        \kldiv(q(M_t | H_{<t}) \Vert p(M_t | H_{<t}))
        \\
        &= 
        \doubleE_q [\ln q(M_t | H_{<t}) - \ln p(M_t | H_{<t})]
        \\
        &=
        \doubleE_q [\ln q(M_t | H_{<t}) - \ln p(M_t, H_{<t}) + \ln p(H_{<t})]
        \\
        &= 
        \ln p(H_{<t}) 
        +
        \doubleE_q [\ln q(M_t | H_{<t}) - \ln p(M_t, H_{<t})]
        \\
        &= 
        \ln p(H_{<t}) 
        +
        \doubleE_q [\ln q(M_t | H_{<t}) - \ln p(H_{<t} | M_t) - \ln p(M_t)]
        \\
        &= 
        \ln p(H_{<t}) 
        -
        \doubleE_q [\ln p(H_{<t} | M_t)] 
        + 
        \kldiv[q(M_t | H_{<t}) \Vert p(M_t)]
        \\
        &= 
        \ln p(H_{<t}) 
        -
        \scriptL_\scriptZ(b^q_t)
    \end{align}
    where the right term is another ELBO,
    \begin{align}
        \scriptL_\scriptZ(b^q_{t+1}) &= 
        \sum_{i=1}^t
        \doubleE_{
            b^q_{t+1}
        }
        [
            \ln p(S_{t+1} | S_t, A_t, M_t) 
            + \ln p(R_t | S_t, A_t, M_t)
            \nonumber
            \\
            & \hspace{8em}
            + \ln \pi(A_t | S_t, M_t)
        ]
        - 
        \kldiv(b^q_{t+1} \Vert b_0),
    \end{align}
    which can be read out from \myeqref{vb:eq:rl_history_latents}.
    Finally, observe $\max_q \Omega^t_\scriptZ = \max_q \scriptL^\scriptZ(q)$ as the log-evidence is a constant.
\end{numproof}

\section{Control as Inference Details}

\subsection{On Determinstic Rewards in Control-as-Inference} \label{vb:ap:delta_rewards_cai}

The probabilistic graphical model we consider in the main text (Figure~\ref{vb:fig:cai_belief}) assumes a deterministic reward function to prevent maximizing behavior over noise.
Although the Control-as-Inference \cite{levine2018reinforcement, ziebart_modeling_2010} does not prevent us from working with stochastic rewards, we should define the outcome variables $\scriptO_t$ to only depend on the mean reward for parity with reinforcement learning objectives.

To see why, assume an MDP setting where we have $\mu_{R_t}$ the mean of the reward function and a noisy realization of $R_t$.
The inverse problem posed by $p(\scriptO_t | A_t, S_t)$ can be expressed as $\frac{1}{\scriptZ} \int p(\scriptO_t | R_t) p(R_t | S_t, A_t)dR_t$ in the stochastic setting, and simply as $\frac{1}{\scriptZ} \int p(\scriptO_t | R_t) \delta(R_t - \mu_R(S_t, A_t))dR_t = \frac{1}{\scriptZ} p(\scriptO_t | \mu_R(S_t, A_t)) = \frac{1}{\scriptZ} \exp \mu_R(S_t, A_t)$ in the deterministic setting (with $\scriptZ$ the normalizing constant).
Suppose the noisy rewards were Gaussian distributed $R_t \sim \scriptN(R_t, \mu_{R_t}, \sigma_R^2)$, then the integration over $R_t$ is simply given by the moment generating function $(\scriptO_t | A_t, S_t) = \frac{1}{\scriptZ}  \exp (\mu_{R_t} + \sigma_R^2 / 2)$. 
Intuitively, when evaluating $p(\scriptO_t | A_t)$ over the entire action space $\scriptA | S_t$, the role of the exponential function is to then smoothly shift probability density to the actions that have higher rewards, and shrink density from actions with low rewards. 
In other words, this example shows a softmax (Boltzmann) policy based on the rewards.
Thus, naively using noisy rewards in the Control-as-Inference framework leads to behavior that seeks out noisy rewards, and using a mean dependency for $\scriptO_t$ does not.

This problem can be easily remedied though, namely by using a model to learn the average rewards.
In our setting, this means that SMC-planning should use the average-reward prediction (and not samples) to compute importance-sampling weights, as discussed in Section~\ref{vb:sec:belief_smc}.
Furthermore, we should also use value estimators and learning in the outer-loop such that this is consistent with the expected returns, which is almost always the case, except in the subtle case of some distributional settings \cite{distributional_bellamare_2017}.

% ...


\section{Experiment Details} \label{vb:ap:varibased_setup}


\subsection{Environments}  \label{vb:ap:setup:envs}
For the simple discrete environment setup, we used our own implementation of a gridworld.
This simple gridworld is almost equivalent to the implementation by \citet{zintgraf_varibad_2020}, where upon resetting the environment, a random goal and start tile are drawn in an open-grid. 
The agent must then explore each tile to find the hidden goal-tile, the agent can explore over multiple episodes each with a strict timelimit.

For the simple continuous control, we used a continuous function optimization task similar to Bayesian optimization methods \cite{chen_learning_2017}.
This environment is stateless, similar to the Bandit setting, however the belief-state is of course maintained over multiple interactions (episodes).
Upon resetting the environment, we generate a new function for the agent to optimize.



\begin{figure}[th]
    \centering
    \includegraphics[width=0.32\linewidth]{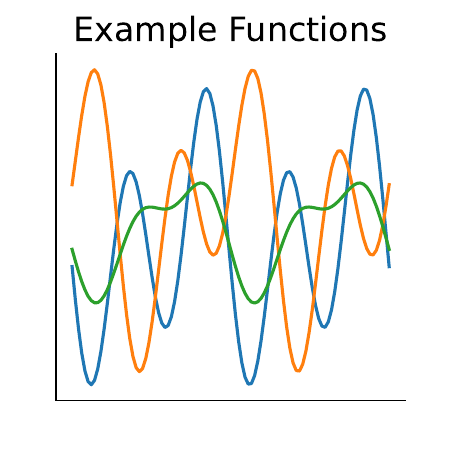}
    \includegraphics[width=0.32\linewidth]{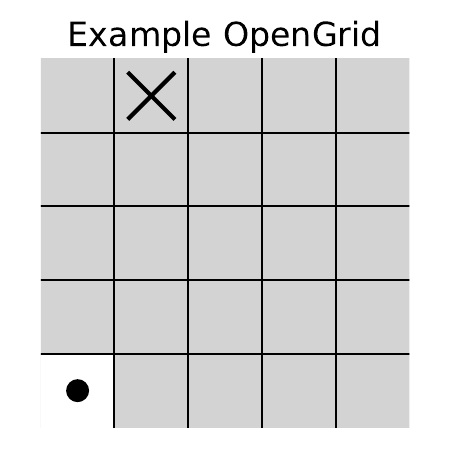}
    \caption{Visualization of environments. Left: continuous function optimization task, where each color corresponds to a distinct function that the agent has to learn to optimize from sequential interactions. Right: discrete opengrid task where the agent needs to find the goal tile $\times$ over multiple episodes starting from the dot. 
    }
    \label{vb:fig:environments}
\end{figure}


Continuous Function optimization environment details (ours):
\begin{itemize}
    \item The observation is the output value of the $n$-dimensional function given the previous input.
    \item The reward is the average over all function dimensions for the given input, $r_t = \frac{1}{D} \sum_{i=1}^D f_i(x_{t,i})$.
    \item The agent can interact with the function for 20 total steps. Although the problem is stationary, it means that the belief-state horizon is 20 steps.
    \item Upon a reset, new functions $\{f_i\}_{i=1}^D$ are generated i.i.d. $\forall f_i$, by randomly sampling the parameters of a $n=3$ truncated Fourier series. For this, we uniformly sample the amplitudes $a \in [0.1, 1.1]$, phase $\phi \in [0, \pi]$, and an input shift $c \in [0, \pi]$, for each component-wave, such that $f(x) = a_0 + \sum^{n-1}_{i=1} a_i \cos (2\pi (x - c) - \phi)$.
\end{itemize}

Discrete GridWorld environment details (ours):
\begin{itemize}
    \item The observation is a $n\times n$ image with 1 channel of values between $[0, 1]$, where $0$ means an empty tile and $1$ means the agent's current position.
    \item The action space is a discrete choice of moving up, down, left, right, or staying still. If the agent moves against a wall, it remains on the same tile.
    \item The reward is zero everywhere, except for a reward of $\frac{1}{\text{time}}$ when the agent is on the goal tile.
    \item The agent terminates after 10 environment steps. The environment is repeated for 6 episodes in total before resetting.
    \item Upon a reset, the goal and start tile are randomly initialized somewhere in the grid.
\end{itemize}

\subsection{Hardware Requirements} \label{vb:ap:hardware}
All experiments were run on a GPU cluster with a mix of NVIDIA Tesla V100-SXM2 32GB, NVIDIA A40 48GB, and A100 80GB GPU cards. 
Each run (random seed/ repetition) required only a few CPU cores (2 logical cores) with a low memory budget (e.g., 4GB). 
We also tested locally on a NVIDIA GeForce RTX 4080, for which most agents could be trained in a few hours. 
Thus, our implementation is readily applicable on consumer grade hardware.

\begin{figure}
    \centering
    \begin{minipage}[t]{0.5\textwidth}
        \input{varibased/figures/varibased_em_pseudocode}
    \end{minipage}
\end{figure}

\subsection{Hyperparameters, Model Training, and Software Versioning}  \label{vb:ap:hyper}

The hyperparameters for our experiments are summarized in Table~\ref{vb:tab:hps}.
Note that all parameter values indicated in sets were run in an exhaustive grid for the ablations. 

Given our final loss function $\scriptL(\theta, \scriptD)$, recall that the E-step is completed for the variational belief $b_{\phi_t}^q$ by doing gradient-descent on the loss to produce $\theta_{(n)}' = \theta_{(n)} - \alpha \nabla \scriptL(\theta)|_{\theta=\theta_{(n)}}$, its  M-step is then performed as $\theta_{(n+1)} \leftarrow \theta_{(n)}'$.
This mixes together the M-step for the policy, whose E-step we already performed during data-generation.
In other words, the E-step for the variational belief, is mixed together with the M-step for the policy/ value network.

We optimize $\scriptL$ with stochastic gradient descent (see the hyperparameter tables) using the AdamW optimizer \cite{loshchilov2019decoupled} with an \( l_2 \) penalty of \(10^{-6}\) and a learning rate of \(3 \cdot 10^{-3}\). Gradients were clipped using two methods, in order: a max absolute value of 1 and a global norm limit of 1.
The overall pseudocode of the outer-loop is provided in Algorithm~\ref{alg:pseudocodeouter}.

The replay buffer was implemented as a uniform circular buffer with size:
\[
\text{max-age of data} \times \text{number of parallel environments} \times \text{number of unroll steps}.
\]
The max-age of data was tuned to fit into reasonable GPU memory.

Additionally, Table~\ref{vb:tab:softwareversions} provides version information for key software packages.
We implemented everything based on Jax~0.4.30 in Python~3.12.4.

\begin{table}[p]
    \centering
    \caption{Experiment hyperparameters. Values in sets were varied as part of our ablations.}
    \label{vb:tab:hps}
    \vspace{0.3em}
    \renewcommand{\arraystretch}{1.1}
    
    \begin{tabular}{|l|c|c|c|}
         \hline
         \textbf{Name} & \textbf{Symbol} & \textbf{Value Function Env} & \textbf{Value Grid} 
         \\
         \hline
         SGD Minibatch size    &    & 1024  & 1024
         \\
         SGD update steps    &    & 32  & 32
         \\
         Unroll length (nr. steps in env)    &    &  64 &  128  
         \\
         Batch-Size (nr. parallel envs)    &    &  32 &  32 
         \\
         (outer-loop) TD-Lambda    &  $\lambda$   &  1.0 &  1.0
         \\
         (outer-loop) Discount    &  $\gamma$   &  0.99 & 0.99 
         \\
         Value Loss Scale    &  $c_v$  &  0.5 &  0.5  
         \\
         Policy Loss Scale    &  $c_\pi$  &  1.0 &  1.0
         \\
         Policy Entropy Loss Scale    &  $c_{ent}$  &  0.0003 &  0.1
         \\
         Loss belief-state burn-in length & & 8 & 12
         \\
         Loss decode window size & & 4 & 6
         \\
         Loss unroll window size & & 4 & 6
         \\
         Belief KL-penalty & & 0.01 & 0.01
         \\
         Belief Entropy Penalty & & $10^{-5}$ & $10^{-5}$
         \\
         \hline
         \textbf{PPO} & & &
         \\
         PPO Policy-Ratio Clipping    &  $\epsilon$  &  0.3 &  0.3
         \\
         PPO Value loss Reward Scale    &   &  10 &  10
         \\
         \hline
         \textbf{SMC} & & &
         \\
         Replay Buffer max-age    &    &  16 & 16 
         \\
         Planner Depth    &  $m$   &  $\{1, 2, 4, 6\}$ &  $\{1, 2, 4, 6\}$ 
         \\
         Number of particles    &  $K$   &  $\{8, 16, 32, 64\}$ &  $\{8, 16, 32, 64\}$ 
         \\
         Resampling period  & $r$  &  2 &  2 
         \\
         Target temperature   &  $T$   &  0.1 &  0.1 
         \\
         \hline
    \end{tabular}
\end{table}

\begin{table}[p]
    \centering
    \caption{Software module versioning that we used for our experiments (also includes default parameter settings).} \label{vb:tab:softwareversions}
    \vspace{0.3em}
    \begin{tabular}{|l|c|}
        \hline
        \textbf{Package} & \textbf{Version} \\ 
        \hline
        brax & 0.10.5 \\
        optax & 0.2.3 \\
        flashbax & 0.1.2 \\
        rlax & 0.1.6 \\
        flax & 0.8.4 \\
        xland-minigrid & 0.9.1 \\
        \hline
    \end{tabular}
\end{table}

\subsection{Neural Network Architectures}

\paragraph{Continuous Function Environment}
\begin{itemize}
    \item Embedding: 3-layer MLP with 128, 64, 32 nodes per layer for the action, observation, and reward separately.
    \item Leaky-ReLU activations followed by LayerNorm.
    \item S5 architecture with 4 layers and Layernorm post-transformation. Other parameters follow the defaults of \citet{lu_s5rl_2023}.
    \item The hidden-state of the S5 is projected to a 32 dimensional Gaussian with independent variances. We use 10 samples for the ELBO through reparametrization to obtain multimodal predictive distributions.
    \item Policy, value, and reward networks all use a 3-layer MLP with 128, 64, 32 nodes per layer.
    \item For the policy, the last layer is linearly projected to parametrize a diagonal multivariate Gaussian squashed via Tanh, as in \citet{haarnoja_sac_2018}.
    \item For the value and reward, the last layer is projected to a scalar to parametrize the mean of a univariate Gaussian.
\end{itemize}

\paragraph{Discrete Grid Environment}
\begin{itemize}
    \item Embedding: 3-layer MLP with 128, 64, 32 nodes per layer for the action and reward separately.
    \item Embedding: 2-layer CNN with 4 channels, a stride of 2, and a 3 by 3 kernel for the grid-observation.
    \item Leaky-ReLU activations followed by LayerNorm.
    \item S5 architecture with 4 layers and Layernorm post-transformation. Other parameters follow the defaults of \citet{lu_s5rl_2023}.
    \item The hidden-state of the S5 is projected to a 32 dimensional Gaussian with independent variances. We use 10 samples for the ELBO through reparametrization to obtain multimodal predictive distributions.
    \item Policy, value, reward, and state networks all use a 3-layer MLP with 128, 64, 32 nodes per layer.
    \item For the policy, the last layer is linearly projected to parametrize the logits.
    \item For the value and reward, the last layer is projected to a scalar to parametrize the mean of a univariate Gaussian.
    \item For the state prediction network, the last layer is projected to independent Bernoulli probabilities for each tile in the observed grid. This probability is a transition probability to predict where the agent moves towards.
\end{itemize}

\end{document}